\documentclass[conference]{IEEEtran}
\usepackage{times}
\usepackage{color, colortbl, xcolor}
\usepackage{multicol}
\usepackage[bookmarks=true,hidelinks]{hyperref}
\usepackage{amsfonts}
\usepackage{amsmath}
\usepackage{dsfont}
\usepackage{multicol}
\usepackage[pdftex]{graphicx}
\usepackage{graphicx}
\usepackage[capitalize,noabbrev,nameinlink]{cleveref}
\crefformat{equation}{(#2#1#3)}
\usepackage[acronyms, shortcuts]{glossaries}
\usepackage[sort,numbers]{natbib}
\usepackage{cases}
\usepackage{xspace}
\usepackage{booktabs}
\usepackage{siunitx}
\usepackage{float}

\usepackage{subfigure}

\glsdisablehyper

\newacronym{ioc}{IOC}{inverse optimal control}
\newacronym{lqr}{LQR}{linear-quadratic regulator}
\newacronym{kkt}{KKT}{Karush–Kuhn–Tucker}
\newacronym{irl}{IRL}{inverse reinforcement learning}
\newacronym{mle}{MLE}{maximum likelihood estimation}
\newacronym[longplural=open-loop Nash equilibria,plural=OLNE]{olne}{OLNE}{open-loop Nash equilibrium}
\newacronym[longplural={partially observable Markov decision processes}]{pomdp}{POMDP}{partially observable Markov decision process}
\newacronym{svo}{SVO}{social value orientation}
\newacronym{ukf}{UKF}{unscented Kalman filter}
\newacronym{ibr}{IBR}{iterated best response}
\newacronym{awgn}{AWGN}{additive white Gaussian noise}
\newacronym{iqr}{IQR}{interquartile range}

\newenvironment{rcases}
  {\left.\begin{aligned}}
  {\end{aligned}\right\rbrace}
\newcommand{\ie}{i.e.\@\xspace}
\newcommand{\eg}{e.g.\@\xspace}

\newcommand{\norm}[1]{\left\lVert#1\right\rVert}
\newcommand{\given}{\mid}
\newcommand{\mbb}{\mathbb}

\newcommand{\mbf}{\mathbf}
\newcommand{\numplayers}{N}
\newcommand{\player}[1]{\text{P$#1$}}
\newcommand{\dynamics}[1]{f_{#1}}
\newcommand{\capdynamics}{\mbf{F}}
\newcommand{\state}{x}
\newcommand{\control}[1]{u^{#1}}
\newcommand{\bstate}{\mathbf{x}}
\newcommand{\bcontrol}[1]{\mathbf{u}^{#1}}
\newcommand{\bcontrolt}{u_t}
\newcommand{\xdim}{n}
\newcommand{\udim}[1]{m^{#1}}
\newcommand{\horizon}{T}
\newcommand{\cost}[1]{J^{#1}}
\newcommand{\runningcost}[1]{g^{#1}}
\newcommand{\costate}[1]{\lambda^{#1}}
\newcommand{\bcostate}[1]{\boldsymbol{\lambda}^{#1}}
\newcommand{\costparams}[1]{\theta^{#1}}
\newcommand{\costparamdim}[1]{k^{#1}}

\newcommand{\observation}{y}
\newcommand{\bobservation}{\mathbf{\observation}}
\newcommand{\noise}{n}
\newcommand{\noisecov}{\Sigma}
\newcommand{\game}{\Gamma}
\newcommand{\kktresidual}{\mathbf{G}}
\newcommand{\indicator}{\ensuremath{\mathbf{1}}\xspace}
\newcommand{\obsmap}{h}
\newcommand{\position}{p}
\newcommand{\speed}{v}
\newcommand{\heading}{\psi}
\newcommand{\acceleration}{a}
\newcommand{\yawrate}{\omega}
\newcommand{\weight}{w}

\newcommand{\goalmap}{\textnormal{d}}
\DeclareGraphicsExtensions{.pdf,.png}

\usepackage{ifthen}
\newboolean{include-notes}
\setboolean{include-notes}{true}

\makeatletter
\def\NAT@spacechar{~}
\makeatother

\usepackage{lipsum}

\newcommand{\vegascale}{0.45}

\newcommand{\example}[1]%
{
\textbf{Running example:}
#1
}

\begin{document}
\title{Inferring Objectives in Continuous Dynamic Games\\ from Noise-Corrupted Partial State Observations}

\author{
\authorblockN{
Lasse Peters\authorrefmark{1},
David Fridovich-Keil\authorrefmark{2},
Vicenç Rubies-Royo\authorrefmark{3},
Claire J. Tomlin\authorrefmark{3} and
Cyrill Stachniss\authorrefmark{1}
}
\authorblockA{
\authorrefmark{1} University of Bonn, Germany\qquad
\authorrefmark{2} University of Texas, Austin, USA\qquad
\authorrefmark{3} University of California, Berkeley, USA\\
Email: \{%
\href{mailto:lasse.peters@igg.uni-bonn.de}{lasse.peters},
\href{mailto:cyrill.stachniss@igg.uni-bonn.de}{cyrill.stachniss}%
\}@igg.uni-bonn.de,
\href{mailto:dfk@utexas.edu}{dfk@utexas.edu},
\{%
\href{mailto:vrubies@berkeley.edu}{vrubies},
\href{mailto:tomlin@berkeley.edu}{tomlin}%
\}@berkeley.edu
}
}

\maketitle

\begin{abstract}
Robots and autonomous systems must interact with one another and their environment to provide high-quality services to their users.
Dynamic game theory provides an expressive theoretical framework for modeling scenarios involving multiple agents with differing objectives interacting over time.
A core challenge when formulating a dynamic game is designing objectives for each agent that capture desired behavior.
In this paper, we propose a method for inferring parametric objective models of multiple agents based on observed interactions.
Our inverse game solver jointly optimizes player objectives and continuous-state estimates by coupling them through Nash equilibrium constraints.
Hence, our method is able to directly maximize the observation likelihood rather than other non-probabilistic surrogate criteria.
Our method does not require full observations of game states or player strategies to identify player objectives.
Instead, it robustly recovers this information from noisy, partial state observations.
As a byproduct of estimating player objectives, our method computes a Nash equilibrium trajectory corresponding to those objectives.
Thus, it is suitable for downstream trajectory forecasting tasks.
We demonstrate our method in several simulated traffic scenarios.
Results show that it reliably estimates player objectives from a short sequence of noise-corrupted partial state observations.
Furthermore, using the estimated objectives, our method makes accurate predictions of each player's trajectory.
\end{abstract}

\IEEEpeerreviewmaketitle

\section{Introduction}
\label{sec:intro}

Most robots use motion planning and optimal control methods to select and execute actions when operating in the real world. Commonly used approaches require specifying the objective to optimize.
In many real-world applications, however, designing optimal control objectives is challenging. 
For example, tuning cost parameters, even in the case of a \ac{lqr}, can be a tedious heuristic process when performed manually. 
As a result, it can be desirable to learn optimal control objectives automatically from demonstrations. 
To this end, researchers have investigated learning from demonstration and \ac{ioc}.
Recent work shows promising results, even for complex problems with large state and observation spaces~\cite{englert2018ijrr,menner2020arxiv}.

\begin{figure}[t]
  \centering
  \includegraphics[scale=\vegascale]{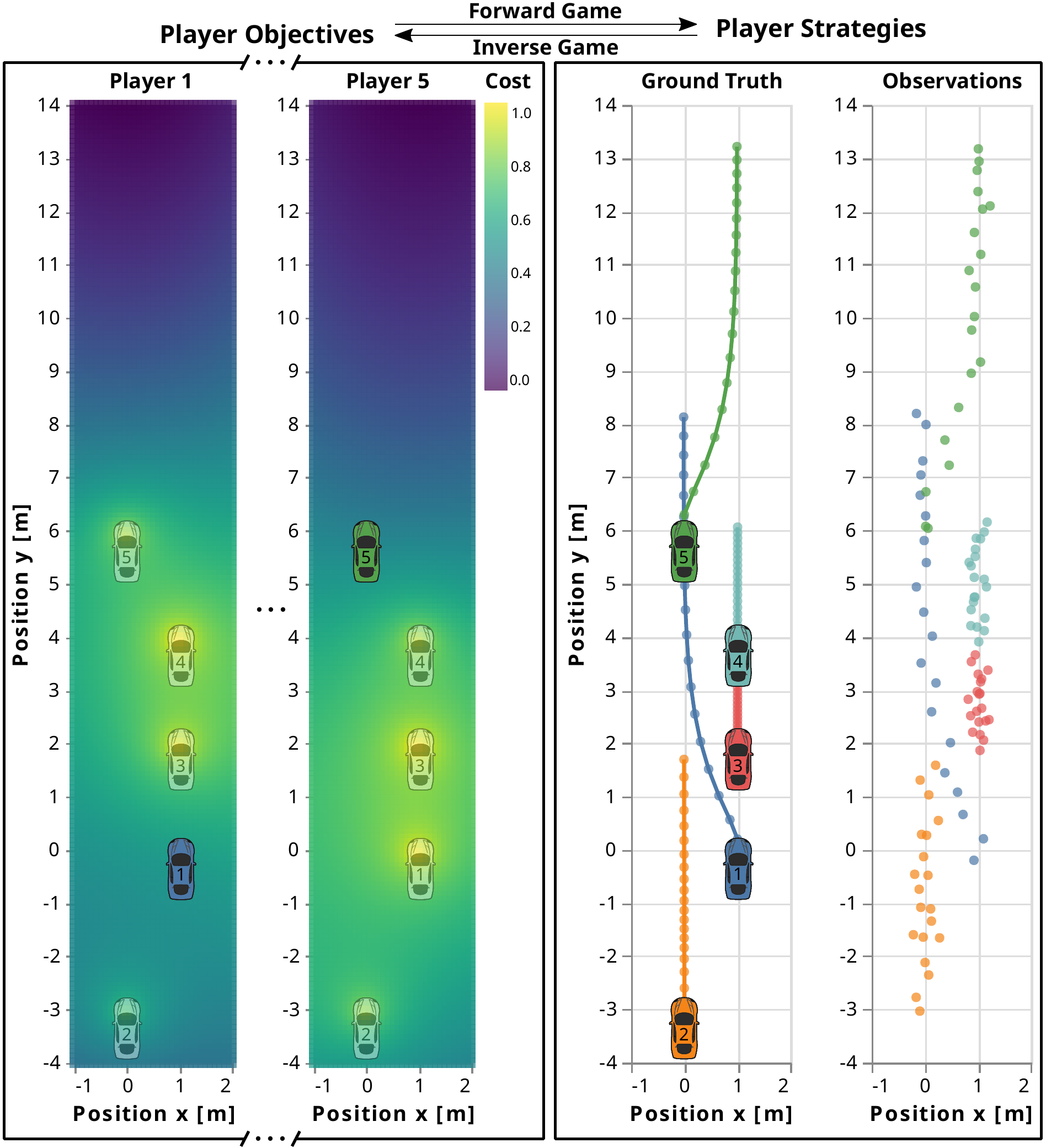}
  \caption{Inverse and forward versions of a dynamic game modeling a 5-player highway driving scenario.
  The solution of the forward problem maps the player objectives (left) to the players' optimal strategies (right).
  Our method solves the inverse problem: it takes noisy, partial state observations of multi-agent interaction as input to recover an objective model for each player that explains the the observed behavior.
  The visualized slice of the cost landscape shows one important aspect of the recovered objective model, namely, each player's preference to keep a safe distance from others.
  The inferred objectives defines an abstract game-theoretic behavior model that can be used to predict player strategies for arbitrary agent configurations.}
  \label{fig:frontfig}
\end{figure}

Optimal control methods, however, are not directly suitable for interactive settings with multiple agents.
For example, consider multiple vehicles engaged in lane changes on a crowded highway.
In this setting, each agent has its own objective that naturally depends upon the behavior of others. 
For instance, agents may wish to maintain a safe distance from others and at the same time travel at a preferred speed.
Thus, their interaction is more accurately characterized as a noncooperative game-theoretic equilibrium rather than as the solution to a joint optimal control problem.
Despite the added complexity of these noncooperative interactions, recent developments enable computationally-efficient solutions to the dynamic games which arise in multi-agent robotic settings~\cite{fridovich2020icra, cleac2020rss, di2019cdc, fridovich2020icra-aiqm}.

There are similar challenges in designing objectives for dynamic games as for single-player optimal control problems.
As in the single-player case, automatic cost learning promises to circumvent this difficulty.
However, cost inference takes on an even more important role in multi-agent settings.
That is, any individual player must also understand the objectives of other players to interact effectively.
In this paper, we study the problem of identifying such cost functions from noisy, partial state observations of multi-agent interactions.
\Cref{fig:frontfig} shows an overview of this problem, commonly referred to as an \emph{inverse} dynamic game.

The main contribution of this paper is a novel, noise-robust technique that identifies unknown parameters of each player's cost function in a noncooperative game based on observed multi-agent interactions.
Recovering these unknown parameters allows us to infer important aspects of each players preferences.
For example, in the highway driving setting of \cref{fig:frontfig} one such aspect is each agent's preference to avoid collisions with other vehicles.
We propose a solution approach to the inverse dynamic game problem which estimates all player's states and control inputs jointly with their unknown objective parameters by coupling them through noncooperative equilibrium constraints.
Through this formulation, our method is can operate seamlessly with noise-corrupted partial state observations.
Based on simulated traffic scenarios, we evaluate our method and provide comparisons to existing methods in a Monte Carlo study.
We show that our method is more robust to incomplete state information and observation noise.
As a result, our method identifies player objectives more reliably, and predicts player trajectories more accurately.

\section{Related Work}
\label{sec:related}

We begin by discussing recent advances in the well-studied area of \ac{ioc}.
While methods from that field address only single-player settings, this body of work exposes many of the important mathematical and algorithmic concepts that appear in games.
We discuss how some of these approaches have been applied in the multi-player setting and emphasize the connections between existing approaches and our contributions.

\subsection{Single-Player Inverse Optimal Control}\label{sec:related-single-player}

The \ac{ioc} problem has been extensively studied since the well-known work of \citet{kalman1964jbe}.
In the context of \ac{irl}, early formulations such as that of \citet{ng2000icml} and maximum-entropy variants~\cite{ziebart2008aaai} have proven successful in treating problems with discrete state and control sets.
In robotic applications, optimal control problems typically involve decision variables in a continuous domain.
Hence, recent work in \ac{ioc} differs from the \ac{irl} literature mentioned above as it is explicitly designed for smooth problems.

One common framework for addressing \ac{ioc} problems with nonlinear dynamics and nonquadratic cost structures is bilevel optimization~\cite{mombaur2010ar,albrecht2011humanoids}.
Here, the outer problem is a least squares or \ac{mle} problem in which demonstrations are matched with a nominal trajectory estimate and decision variables parameterize the objective of the underlying optimal control problem.
The inner problem determines the nominal trajectory estimate as the optimizer of the ``forward'' (i.e., standard) optimal control problem for the outer problem's decision variables.
A key benefit of bilevel \ac{ioc} formulations is that they naturally adapt to settings with noise-corrupted partial state observations~\cite{albrecht2011humanoids}. 

Early bilevel formulations for \ac{ioc} utilize derivative-free optimization schemes to estimate the unknown objective parameters in order to avoid explicit differentiation of the solution to the inner optimal control problem~\cite{mombaur2010ar}.
That is, the inner solver is treated as a black-box mapping from cost parameters to optimal trajectories which is utilized by the outer solver to identify the unknown parameters using a suitable derivative-free method.
While black-box approaches can be simple to implement due to their modularity and lack of reliance on derivative information, they often suffer from a high sampling complexity~\cite{nocedal2006optimizationbook}. 
Since each sample in the context of black-box \ac{ioc} methods amounts to solving a full optimal control problem, such approaches remain intractable for scenarios with large state spaces or additional unknown parameters, such as unknown initial conditions.

Other works instead embed the \ac{kkt} conditions of the inner problem as constraints on the outer problem.
Since these techniques enforce only first-order necessary conditions of optimality, globally optimal observations are unnecessary and locally optimal demonstrations suffice.
Yet, a key computational difficulty of \ac{kkt}-constrained \ac{ioc} formulations is that they yield a nonconvex optimization problem due to decision variables in the outer problem appearing nonlinearly with inner problem variables in \ac{kkt} constraints. 
This occurs even in the relatively benign case of linear-quadratic \ac{ioc}.

In contrast to bilevel optimization formulations where necessary conditions of optimality are embedded as constraints, recent methods \cite{levine2012icml,englert2018ijrr,awasthi2019master,menner2020arxiv} minimize the residual of these conditions directly at the demonstrations.
Since the observed demonstration is assumed to satisfy any constraints of the underlying forward optimal control problem, this method can be formulated as fully unconstrained optimization.
Additionally, these residual formulations yield a \emph{convex} optimization problem if the class of objective functions is convex in the unknown parameters at the demonstration \cite{keshavarz2011isic, englert2018ijrr}. This condition holds in the common setting of linearly-parameterized objective functions.
\citet{levine2012icml} propose a variant of this approach that utilizes quadratic approximations of the reward model around demonstrations to derive optimality residuals in a maximum entropy framework.
\citet{englert2018ijrr} present an extensions of this method do accommodate inequality constraints on states and inputs. 
Much like \ac{kkt}-constrained formulations, these residual methods operate on locally optimal demonstrations.
However, an important limitation of residual methods is that they require observations of full state and input sequences.
More recently, \citet{menner2020arxiv} compared \ac{ioc} techniques based on \ac{kkt} constraints and residuals and demonstrated inferior performance of the latter even in problems with linear dynamics and quadratic target objectives.

Our work takes inspiration from the \ac{kkt}-constraint formulation for single-player \ac{ioc} as discussed by \citet{albrecht2011humanoids} and \citet{menner2020arxiv}.
While these work apply only to single-player settings, we utilize necessary conditions for \acp{olne} to generalize this approach to noncooperative multi-player scenarios.


\subsection{Multi-Player Inverse Dynamic Games}\label{sec:related-multi-player}

Many of the \ac{ioc} techniques discussed above have close analogues in the context of multi-player inverse dynamic games.

As in single-player \ac{ioc}, methods akin to black-box bilinear optimization have also been studied in the context of inverse games \cite{peters2020master,cleac2020arxiv}.
\citet{peters2020master} use a particle-filtering technique for online estimation of human behavior parameters.
This work demonstrates the importance of inferring human behavior parameters for accurate prediction in interactive scenarios.
However, there, inference is limited to a single parameter and the work highlight the challenges associated with scaling this sampling based approach to high-dimensional latent parameter spaces.
\citet{cleac2020arxiv} employ a similar derivative-free filtering technique based on an \acl{ukf}.
While this approach drastically reduces the overall sample complexity, it still relies on exact observations of the state to reduce the required number of solutions to full dynamic games at the inner level.

Another line of research has put forth solution techniques for inverse games that follow from the residual methods outlined in \cref{sec:related-single-player} \cite{kopf2017ifac,rothfuss2017ifac,awasthi2020acc,inga2019arxiv}.
\citet{kopf2017ifac} study a special case of an inverse linear-quadratic game in which the equilibrium feedback strategies of all but one player are known.
This assumption reduces the estimation problem to single-player \ac{ioc} to which the residual methods discussed above can be applied directly.
\citet{rothfuss2017ifac} present a more general approach that does not exploit such special structure but instead minimizes the residual of the first-order necessary conditions for a local \ac{olne}.
\citet{inga2019arxiv} present a variant of this \ac{olne} residual method in a maximum entropy framework, generalizing the single-player \ac{ioc} algorithm proposed by \citet{levine2012icml}.
Recently, \citet{awasthi2020acc} also extended the \ac{olne} residual method of \citet{rothfuss2017ifac} to inverse games with state and input constraints. This approach extends that of \citet{englert2018ijrr} to noncooperative multi-player scenarios.

All of these inverse game \ac{kkt} residual methods share many properties with their single-player counterparts.
In particular, since they rely upon only local equilibrium criteria, they are able to recover player objectives even from local---rather than only global---equilibrium demonstrations. 
However, as in the single-player case, they rely upon observation of both state and input to evaluate the residuals.

In contrast to \ac{kkt} residual methods \cite{rothfuss2017ifac,awasthi2020acc,inga2019arxiv}, we enforce these conditions as constraints on a jointly estimated trajectory, rather than minimizing the residual of these conditions directly at the observation.
Thus, our method can explicitly account for observation noise, partial state observability, and unobserved control inputs.
Furthermore, in contrast to black-box approaches to the inverse dynamic game problem~\cite{peters2020master,cleac2020arxiv}, our method does not require repeated solutions of the underlying forward game.
Moreover, our method returns a full forward game solution in addition to the estimated objective parameters for all players.


\section{Background: Open-Loop Nash Games}
\label{sec:background}

This section offers a concise background on \emph{forward} open-loop Nash games.
In this work, we use the term forward to disambiguate this class of problems from that of learning costs in games (\ie, inverse games).
For a thorough treatment, refer to \citet{basar1998gametheorybook}. 
Note that \acp{olne} differ from noncooperative equilibrium concepts in other information structures including feedback Nash equilibria \cite[Chapter 3]{basar1998gametheorybook}.
Recent algorithms for open-loop games are those by \citet{di2019cdc, cleac2020rss}, and for feedback games we refer to \citet{fridovich2020icra} and \citet{laine2021arxiv}.

An open-loop (infinite) Nash game with $\numplayers$ players is characterized by \emph{state} $\state \in \mbb{R}^\xdim$ and control inputs for each player $\control{i} \in \mbb{R}^{\udim{i}}$ which follow \emph{dynamics} $\state_{t+1} = \dynamics{t}(\state_t, \control{1}_t,\dots,\control{N}_t)$ at each discrete time $t \in [\horizon] := \{1,\dots,\horizon\}$.
Each player has a cost function\footnote{This setup readily extends to the constrained case as in \cite{cleac2020rss, laine2021arxiv}---as our own work does.
We ignore such constraints here for clarity.} $\cost{i} := \sum_{t=1}^\horizon \runningcost{i}_t(\state_t, \control{1}_t, \dots, \control{\numplayers}_t)$, which is implicitly a function of the initial condition $\state_1$ and explicitly of both the control inputs for each player $\bcontrol{i} := (\control{i}_1, \dots, \control{i}_\horizon)$ and the state trajectory $\bstate := (\state_1, \dots, \state_\horizon)$.
The tuple of initial state, joint dynamics, and player objectives which fully characterizes a game is denoted $\game := \left(\state_1, \dynamics{}, \{\cost{i}\}_{i \in [\numplayers]}\right)$ throughout this work.

Given a sequence of control inputs for all players $\bcontrol{} := (\bcontrol{1}, \ldots, \bcontrol{N})$ the states are determined by the dynamics and initial condition.
Note that for clarity we use bold variables to indicate aggregation over time and omit player indices to further aggregate a quantity over all players.
Hence, for shorthand, we will overload cost notation to define $\cost{i}(\bcontrol{}; \state_1) \equiv \cost{i}(\bcontrol{1}, \dots, \bcontrol{\numplayers}; \state_1) \equiv \cost{i}(\bstate, \bcontrol{1}, \dots, \bcontrol{\numplayers})$.

Nash equilibria are solutions to the coupled optimization problems, one for each player \player{i}:
\begin{subequations}
\label{eqn:nash}
\begin{numcases}{\forall i \in [\numplayers]}
    \label{eqn:coupled_optimization_problems}
    \min_{\bstate, \bcontrol{i}}~\cost{i}(\bcontrol{}; \state_1)\\
    \label{eqn:dynamics_constraint_t}
    \textrm{s.t.}~\state_{t+1} = \dynamics{t}(\state_t, \control{1}_t, \dots, \control{\numplayers}_t), \forall t \in [T-1].
\end{numcases}
\end{subequations}

Nash equilibrium strategies $\bcontrol{*} := (\bcontrol{1*}, \dots, \bcontrol{\numplayers*})$ satisfy the inequality $\cost{1}(\bcontrol{1}, \bcontrol{2*}, \dots, \bcontrol{\numplayers*}; \state_1) \ge \cost{1}(\bcontrol{*}; \state_1)$ for the first player (\player{1}) and likewise for all other players.
Intuitively, at equilibrium no player wishes to unilaterally deviate from their respective strategy $\bcontrol{i*}$.
Note that this solution concept differs from a formulation as joint optimal control problem.
In particular, players' objectives may conflict in which case the resulting equilibrium is \emph{noncooperative}.


\example{
To make these concepts concrete, we introduce the following running example.
Consider $N=2$ vehicles avoiding collision.
Each vehicle has its own state $\state^i$ such that the global game state is concatenated as $\state = (\state^1, \state^2)$.
Further, each vehicle follows unicycle dynamics at time discretization $\Delta t$:
\begin{equation}
\label{eqn:runexp-dynamics}
\state_{t+1}^i =
    \begin{cases}
    \textnormal{\emph{(x-position)}}~\position_{x,t+1}^i &= \position_{x,t}^i + \Delta t ~\speed_t^i \cos \heading_t^i\\
    \textnormal{\emph{(y-position)}}~\hfill\position_{y,t+1}^i &= \position_{y,t}^i + \Delta t ~\speed_t^i \sin \heading_t^i\\
    \textnormal{\emph{(heading)}}~\hfill\heading_{t+1}^i &= \heading_t^i + \Delta t ~\yawrate_t^i\\
    \textnormal{\emph{(speed)}}~\quad\hfill\speed_{t+1}^i &= \speed_t^i + \Delta t ~\acceleration_t^i,
    \end{cases}
\end{equation}
where $\control{i}_t = (\yawrate_t, \acceleration_t)$ is the yaw rate and longitudinal acceleration.
Finally, each player's objective is characterized by a running cost $\runningcost{i}_t$ defined as a combination of multiple basis functions:
\begin{subequations}
\label{eqn:runexp-cost}
\begin{numcases}{\runningcost{i}_t = \sum_{j = 1}^5 \weight^i_j \runningcost{i}_{j,t}}
    \runningcost{i}_{1,t} = \indicator(t \ge T - t_{\textnormal{goal}}) \goalmap(\state_t^i, \state_{\textnormal{goal}}^i)  \label{eqn:runexp-goal-cost}\\
    \runningcost{i}_{2,t} = -\log(\|\position_i - \position_{-i}\|_2^2)\label{eqn:runexp-prox-cost}\\
    g_{3,t}^i = (v^i)^2\label{eqn:runexp-vel-cost}\\
    \runningcost{i}_{4,t} = (\yawrate_t^i)^2 \label{eqn:runexp-yawrate-cost}\\
    \runningcost{i}_{5,t}  = (\acceleration_t^i)^2 \label{eqn:runexp-accel-cost},
\end{numcases}
\end{subequations}
where $\weight_j^i \in \mbb{R}_+$ are non-negative weights for each cost component, $\position_i$ and $\position_{-i}$ denote the position of player $\textrm{P}i$ and its opponent, and $\goalmap(\cdot, \cdot)$ is a distance mapping.
For this example we choose $\goalmap(\state_t^i, \state^i_{\textnormal{goal}}) = \|\position_{t}^i - \position_{\textnormal{goal}}^i\|_2^2$ to compute squared distance from a goal position.
The basis functions encode the following aspects of each player's preferences:
\begin{enumerate}
    \item be close to the goal state in the last $t_{\textnormal{goal}}$ time steps (\ref{eqn:runexp-goal-cost}),
    \item avoid close proximity to the other vehicle (\ref{eqn:runexp-prox-cost}),
    \item avoid high speed (\ref{eqn:runexp-vel-cost}) and large control effort (\ref{eqn:runexp-yawrate-cost}, \ref{eqn:runexp-accel-cost}).
\end{enumerate}
This game is inherently noncooperative since players must compete to reach their own goals safely and efficiently:
No player wishes to deviate from a direct path to the goal, yet all players also wish to avoid collision.
Hence, they must negotiate these conflicting objectives and thereby find an equilibrium of the underlying game.
Note that the cost structure in \cref{eqn:runexp-cost} can also be used to encode more complex problems such as the highway driving scenario depicted in \cref{fig:frontfig}.
For clarity, we limit discussion to a simplified 2-player scenario in this running example, and present a 5-player example later in the paper.
}

\section{Problem Formulation}\label{sec:inverse_problem}

A Nash game requires finding optimal strategies for each player, given their objectives.
In contrast, this work is concerned with the inverse problem that requires finding players' objectives for which the observed behavior is a Nash equilibrium.
In short, it seeks an answer to the question: \emph{Which player objectives explain the observed interaction?}

We cast this question as an estimation problem.
To that end, we assume that each player's cost function is parameterized by a vector $\costparams{i} \in \mbb{R}^{\costparamdim{i}}$, \ie, $\cost{i}(\cdot; \costparams{i}) \equiv \sum_{t=1}^\horizon \runningcost{i}_t(\state_t, \control{1}_t, \dots, \control{\numplayers}_t; \costparams{i})$. 

This formulation includes arbitrary smooth and potentially nonlinear parameterizations. Hence, a player's objective may also be parameterized by a differentiable function approximator such as an artificial neural network.
While such a parameterization is very flexible, it may also reduce the interpretability of the resulting parameters.
The parameterization of player objectives may further be designed to exploit domain knowledge for a specific application.

\example{
For clarity of presentation, our running example throughout this paper considers a linear parameterization $\costparams{i} = (\weight^i_1, \dots, \weight^i_5)$ which weights individual cost functions from \cref{eqn:runexp-cost} that comprise the overall objective for each player.
As we will see in \cref{sec:experiments}, this parametrization is able to capture a variety of interactive traffic scenarios.}

Thus equipped, we seek to estimate those parameter values that maximize the likelihood of a given sequence of partial state observations $\bobservation := (\observation_1, \ldots, \observation_T)$ for the induced parametric family of games $\game(\costparams{}) = \left(\state_1, \dynamics{}, \{\cost{(i)}(\,\cdot\,; \costparams{(i)})\}_{i \in [\numplayers]}\right)$:  
\begin{subequations}
\label{eqn:inverse_problem}
\begin{align}
\label{eqn:inverse_objective}
    \max_{\costparams{}, \bstate, \bcontrol{}} \quad& p(\bobservation \given \bstate, \bcontrol{})\\
\label{eqn:inverse_olne_constraint}
    \textrm{s.t.} \quad& (\bstate, \bcontrol{}) \text{ is an OLNE of } \game(\costparams{})\\
\label{eqn:inverse_dynamics_constraints}
                       & (\bstate, \bcontrol{}) \text{ is dynamically feasible under } f,
\end{align}
\end{subequations}
where, $\costparams{}$ is the vector of aggregated parameters over all players, \ie, $\costparams{} := (\costparams{1}, \ldots, \costparams{\numplayers})$, and $p(\bobservation\given \bstate, \bcontrol{})$ denotes a known observation likelihood model. 

In the simplest case, the inverse planner receives an exact observation of the full state and input sequence and the observation model is a Dirac delta function.
In general, however, the observation model $p(\bobservation \given \bstate, \bcontrol{})$ allows modelling noise-corrupted partial state observations.
Thus, our formulation is amenable to more realistic scenarios and real sensors, for example, range/bearing measurements from a LiDAR.

In summary, the above formulation of the inverse dynamic game problem attempts a \emph{joint} estimation of states, control inputs, and player objectives by tightly coupling them through Nash equilibrium constraints.
Note that this is an important difference to existing formulations \cite{awasthi2020acc, rothfuss2017ifac} which treat these estimation problems separately and do not exploit the strong Nash priors which couple them.
We discuss these methods in further detail below and compare to them as a baseline.

\section{Our Approach} 
\label{sec:solution}
This section describes our main contribution: a novel solution technique for identifying objective parameters of players in a continuous game.
Our formulation is directly expressed in the standard format of a constrained optimization problem.
That is, our method yields a mathematical program which can be encoded using well-established modeling languages (\eg, CasADi~\cite{andersson2019mpc}, JuMP~\cite{dunning2017sirev}, and YALMIP~\cite{lofberg2004icra}) and solved by a number of off-the-shelf methods (\eg, IPOPT~\cite{wachter2006jmp},  KNITRO~\cite{byrd2006lsno}, and SNOPT~\cite{gill2005sirev}).

%

\subsection{Encoding Nash Equilibrium Constraints}
A key challenge to solving the estimation problem in \cref{eqn:inverse_problem} is posed by the requirement to encode the equilibrium constraint in \cref{eqn:inverse_olne_constraint} in order to couple the estimates of game trajectory~$(\bstate, \bcontrol{})$ and objective parameters $\costparams{}$.
In this work, akin to the bilevel optimization approach to single-player \ac{ioc} of \citet{albrecht2011humanoids}, we encode this forward optimality constraint via the corresponding first-order necessary conditions.
For an \ac{olne}, the first-order necessary conditions are given by the union of the individual players' \ac{kkt} conditions, \ie,
\begin{align}
    \label{eqn:forward_kkt_conditions}
    \kktresidual(\bstate, \bcontrol{}, \bcostate{}) := \begin{bmatrix}
    \begin{rcases}
        \nabla_\bstate \cost{i} + \bcostate{i\top} \nabla_\bstate \capdynamics(\bstate, \bcontrol{})\\
        \nabla_{\bcontrol{i}} \cost{i} + \bcostate{i\top} \nabla_{\bcontrol{i}} \capdynamics(\bstate, \bcontrol{})
    \end{rcases}
    \text{$\forall i \in [\numplayers]$}
    \\
    \capdynamics(\bstate, \bcontrol{})
    \end{bmatrix}
    = \mathbf{0}.
\end{align}
\noindent
The first two blocks of this equation are repeated for all players~\player{i} and $\capdynamics(\bstate, \bcontrol{})$ collects the dynamics constraint error from \cref{eqn:dynamics_constraint_t} with  $t^{\textnormal{th}}$ block of $\state_{t+1} - \dynamics{t}(\state_t, \control{1}_t, \dots, \control{\numplayers}_t)$.
Here, we introduce costates $\bcostate{i} := (\costate{i}_1, \dots, \costate{i}_{\horizon-1})$ for all players, where $\costate{i}_t \in \mbb{R}^\xdim$ is the Lagrange multiplier associated with the constraint between decision variables at time step $t$ and $t+1$ in \cref{eqn:dynamics_constraint_t}.


Incorporating \cref{eqn:forward_kkt_conditions} as constraints, we cast the inverse dynamic game problem of \cref{eqn:inverse_problem} as
\begin{subequations}
\label{eqn:inverse_approach}
\begin{align}
    \max_{\costparams{}, \bstate, \bcontrol{}, \bcostate{}} \quad& p(\bobservation \given \bstate, \bcontrol{})\\
\label{eqn:inverse_kkt_constraints}
    \textrm{s.t.} \quad& \kktresidual(\bstate, \bcontrol{}, \bcostate{}; \costparams{})  = \mathbf{0}.
\end{align}
\end{subequations}
Here, the costates $\bcostate{}$ of \cref{eqn:forward_kkt_conditions} appear as \emph{additional primal decision variables}.
Further, $\kktresidual(\bstate, \bcontrol{}, \bcostate{}; \costparams{})$ is the \ac{kkt} residual from \cref{eqn:forward_kkt_conditions}, with added explicit dependency on the cost parameters $\costparams{}$.
In practice, it can also be important to regularize or otherwise constrain parameters.

\example{j
To ensure that the problem remains well-defined, we constrain all parameters to sum to unity for all players, and we also require positive control cost weights, \ie, $\costparams{i}_{4, 5} > \epsilon \ge 0$.
}

Note that \cref{eqn:inverse_kkt_constraints} does not explicitly depend upon observations~$\bobservation$ but instead utilizes the trajectory~$(\bstate, \bcontrol{})$ which we optimize simultaneously to maximize observation likelihood.
Thus, our method does not rely on complete observation of states, or even inputs.
Rather, we reconstruct this missing information by exploiting knowledge of dynamics and objective model structure.

Finally, we also note that our method applies coherently when there are multiple observed trajectories; our development here treats the single-trajectory observation case for clarity.

\subsection{Structure of Constraints}
Consider the $t^\textnormal{th}$ term in the first block of $\kktresidual$ in \cref{eqn:forward_kkt_conditions}
\begin{subequations}
\begin{align}
    0 &= \nabla_{\state_t} \cost{i}(\bstate, \bcontrol{}; \costparams{i}) + \bcostate{i\top} \nabla_{\state_t} \capdynamics(\bstate, \bcontrol{})\\ 
    &= \nabla_{\state_t} \runningcost{i}_t(\state_t, \bcontrolt; \costparams{i}) + \costate{i}_{t-1} - \costate{i\top}_t \nabla_{\state_t} \dynamics{t}(\state_t, \bcontrolt),
\end{align}
\end{subequations}
with aggregated player inputs~$\bcontrolt = (\control{1}_t, \dots, \control{\numplayers}_t)$.
The right hand side contains two potential sources of nonlinearities that may render \cref{eqn:inverse_approach} nonconvex.
First, for any non-trivial player objectives, the gradient of the running cost~$\nabla_{\state_t}\runningcost{}$ must couple state~$\state_t$ and inputs~$\bcontrolt$ with parameters~$\costparams{}$.
Second, costate~$\costate{i}_t$ multiplies the potentially state-dependent Jacobian of dynamics~$\nabla_{\state_t} \dynamics{t}$.

When the dynamics are affine, the Jacobian is a constant and the latter term remains linear in decision variables $\costate{i}_t$ and $\state_t$.
Furthermore, in a forward dynamic game, the parameter vector $\costparams{}$ is given in the problem description and thus also the former nonlinearity vanishes if player objectives are quadratic and include no mixed terms in states and inputs.
As a result of this structure, linear-quadratic \ac{olne} problems admit an analytic solution \cite[Chapter 6]{basar1998gametheorybook} and only more complex problems require the use of iterative solution techniques, \eg a Newton method as employed by \citet{cleac2020rss}.

In an inverse game, however, objective parameters $\costparams{}$ necessarily appear as decision variables.
Since our method additionally estimates the game trajectory $(\bstate, \bcontrol{})$ to account for noise-corrupted partial state observations, the equilibrium constraints in \cref{eqn:inverse_kkt_constraints} remain at least bilinear even for linear-quadratic games and the optimization problem is inevitably nonconvex.
Therefore, our approach inherently relies on an iterative method to identify solutions of \cref{eqn:inverse_approach} and the ability to solve this problem can depend on suitable initialization of the decision variables.

To this end, we leverage the observation sequence $\bobservation$ to initialize the decision variables $\bstate$ and $\bcontrol{}$ by solving a relaxed version of \cref{eqn:inverse_approach} without equilibrium constraints.
That is, we compute the initialization of the state-input trajectory as the solution of
\begin{subequations}
\label{eqn:pre_solve}
\begin{align}
\label{eqn:pre_solve_objective}
\tilde\bstate, \tilde{\bcontrol{}} := \arg\max_{\bstate, \bcontrol{}} \quad & p(\bobservation \given \bstate, \bcontrol{})\\
\label{eqn:pre_solve_constraint}
\textrm{s.t.}  \quad& \capdynamics(\bstate, \bcontrol{}) = \mbf{0}.
\end{align}
\end{subequations}
This pre-solve step can be interpreted as sequentially activating the different components of the \ac{kkt} constraints in \cref{eqn:inverse_kkt_constraints}.
First, we enforce only dynamics constraints to recover a trajectory that maximizes observation likelihood while remaining dynamically feasible, regardless of Nash equilibrium constraints.
Subsequently, we activate the Nash equilibrium constraints encoded by the first two blocks of $\kktresidual$ and solve the full problem in \cref{eqn:inverse_approach}.
Note that in this second step, the state $\bstate$ and input $\bcontrol{}$ still remain decision variables and thus the game trajectory is further refined during optimization of the objective parameters~$\costparams{}$.


\subsection{Maximum Likelihood Objective}

A common yet effective class of observation models assumes \acl{awgn}. 
In this case, each observation depends only upon the state and input at the current time step, \ie,
\begin{align}\label{eqn:awgn_observation}
    \observation_t = \obsmap_t(\state_t, \control{1}_t, \ldots, \control{\numplayers}_t) + \noise_t, 
\end{align}
\noindent
where $\obsmap_t$ is a deterministic mapping from the current state and input to the \emph{expected} observation, and $\noise_t$ is a zero-mean white noise-process, \ie, $\noise_t\sim\mathcal{N}(0, \noisecov_t)$.
For this class of observation models the inverse \ac{olne} problem can be equivalently treated as a constrained nonlinear least-squares problem.

\example{
We presume isotropic \acl{awgn} and minimize the corresponding negative log-likelihood objective $\sum_t \|\observation_t - \obsmap_t(\state_t)\|_2^2$ in \cref{eqn:inverse_objective}.
In summary, the inverse problem to the collision avoidance game entails the following task:
Find those weights to the basis functions in \cref{eqn:runexp-cost} for which the corresponding game solution generates expected observations near the observed data.
}

\section{Experiments}
\label{sec:experiments}

This section analyzes the performance of the proposed inverse game solution approach and compares it to a state-of-the-art baseline in a Monte Carlo study.

\subsection{Baseline: Minimizing KKT Residuals}
\label{sec:baseline}

We use as a baseline the \ac{kkt} residual approach presented in \citet{rothfuss2017ifac} and \citet{awasthi2020acc}.
Other approaches exist as described in \cref{sec:related-multi-player}, \eg, \cite{ cleac2020arxiv,peters2020master}, however these black-box approaches do not utilize derivative information.
Algorithmic differences are sufficiently extensive that they render a direct comparison difficult to interpret.

Like our method, the \ac{kkt} residual approach uses the first-order necessary conditions in \cref{eqn:forward_kkt_conditions} to encode forward optimality.
However, it does not jointly optimize a trajectory estimate for the problem.
Instead, these method assume access to a \emph{preset} trajectory along which they minimize the violation of the optimality constraint.
That is, the baseline solves
\begin{align}
\label{eqn:kkt_residual_baseline}
    \min_{\costparams{}, \bcostate{}} \norm{\kktresidual(\tilde\bstate, \tilde{\bcontrol{}}, \bcostate{}; \costparams{})}_2^2,
\end{align}
where $\tilde\bstate$ and $\tilde{\bcontrol{}}$ are assumed to be given as part of the observation.
Thus, only the objective parameters $\costparams{}$ and the costates $\bcostate{}$ are decision variables in the problem.

In scenarios with incomplete information due to unobserved inputs, noise, or partial state observations, the solution to the optimization problem in \cref{eqn:kkt_residual_baseline} is not meaningful and not always well-defined.
Instead, the state-input trajectory must first be estimated from the observation sequence $\bobservation$ in order to evaluate the constraint residual. 
To this end, we extend the technique of \cite{rothfuss2017ifac, awasthi2020acc} with a pre-processing step to estimate the state-input trajectory as the solution of \cref{eqn:pre_solve}.
That is, we recover the dynamically feasible state-input sequence that maximizes the likelihood of the observation.
This baseline can be thought of as a sequential, loosely coupled version of our approach.
Instead of optimizing a trajectory estimate jointly with the objective parameters $\costparams{}$, they are estimated one at a time in a two-stage procedure.

\subsection{Experimental Setup}

We implement our proposed approach as well as the \ac{kkt} residual baseline \cite{rothfuss2017ifac} in Julia \cite{bezanson2017sirev} using the algebraic modeling language JuMP \cite{dunning2017sirev}.
Due to the abstraction provided by the modeling language our implementation is agnostic to the algorithm used to solve the synthesized problem description.
In this work, we use the open source COIN-OR IPOPT algorithm \cite{wachter2006jmp}.
The source code is publicly available at \href{https://github.com/PRBonn/PartiallyObservedInverseGames.jl}{https://github.com/PRBonn/PartiallyObservedInverseGames.jl}.

To compare robustness and performance of our method with the baseline, we perform a Monte Carlo study.
For this purpose, we fix a set of cost weights in \cref{eqn:runexp-cost} for each player, find corresponding
\ac{olne} trajectories as roots of \cref{eqn:forward_kkt_conditions} using the well-known \ac{ibr} algorithm \cite{wang2019dars}, and
corrupt them with \acl{awgn} as described in \cref{eqn:awgn_observation}.
We generate 40 random observation sequences at each of 22 different levels of isotropic observation noise.
For each of the resulting 880 observation sequences we run both our method and the baseline to recover estimates of weights~$w^i_j, j \in \{1,\dots,5\}$ for each player. 
Note that all methods infer objective parameters from observation of a single trajectory.
That is, each estimate in the Monte Carlo study relies only upon \SI{25}{\second} of interaction history of a single scenario instead of batches of multiple demonstrations.
This evaluation setup is designed to benchmark the methods in a realistic setting where an estimator typically cannot observe the same scene multiple times.

\subsection{Simulation Experiments}
\label{sec:results}
\subsubsection{2-Player Running Example}

First, we evaluate our method and the residual baseline in a Monte Carlo study using the running example of collision-avoidance with $\numplayers = 2$ players.
This experiment aims to demonstrate the performance gap of both methods in a conceptually simple and more easily interpretable scenario.

\begin{figure}
    \centering
    \subfigure[Parameter estimation\label{fig:runexmp-estimator_parameter_error}]{
    \includegraphics[scale=\vegascale]{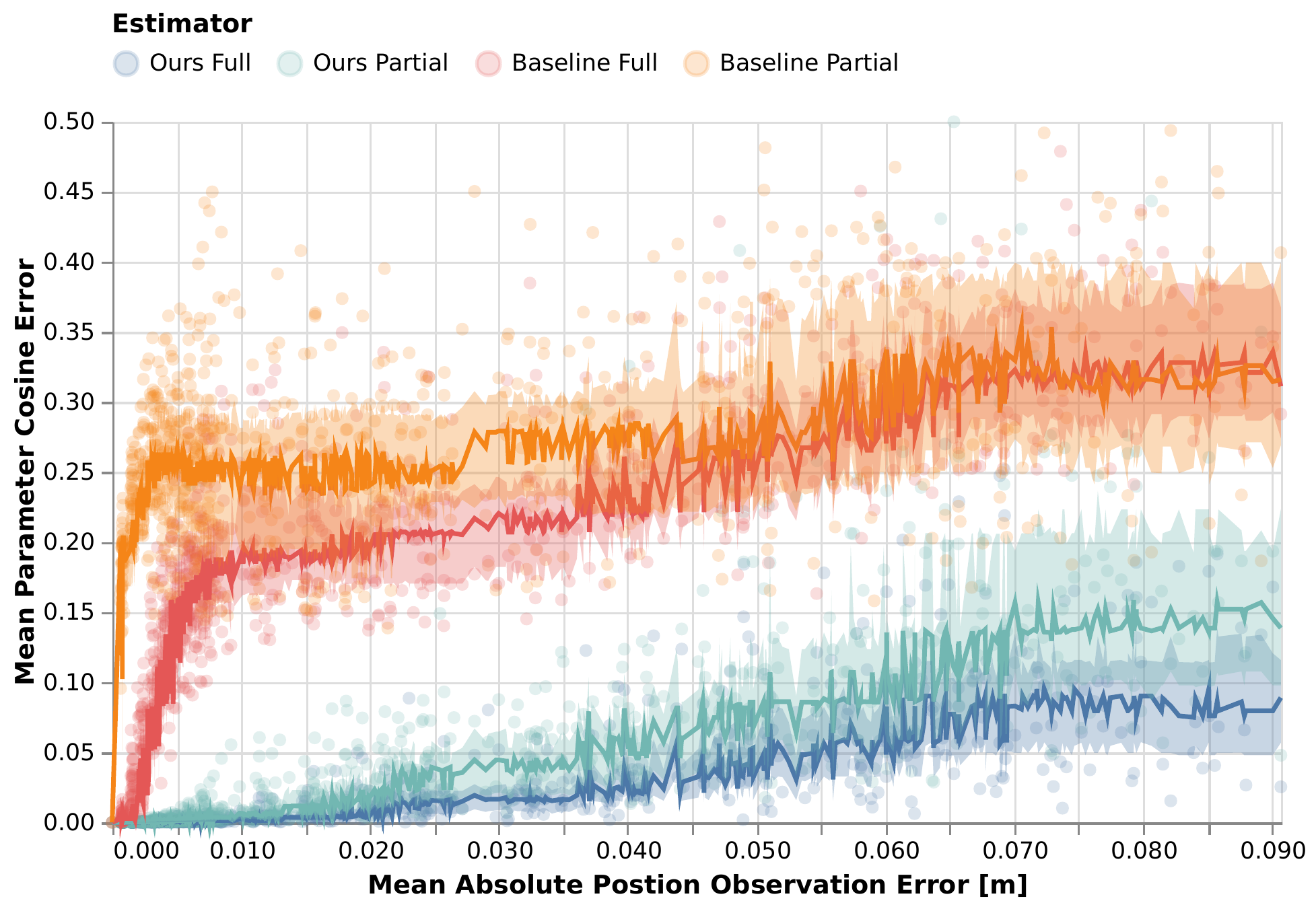}
    }
    \hfill
    \subfigure[Position prediction\label{fig:runexmp-estimator_position_error}]{
    \includegraphics[scale=\vegascale]{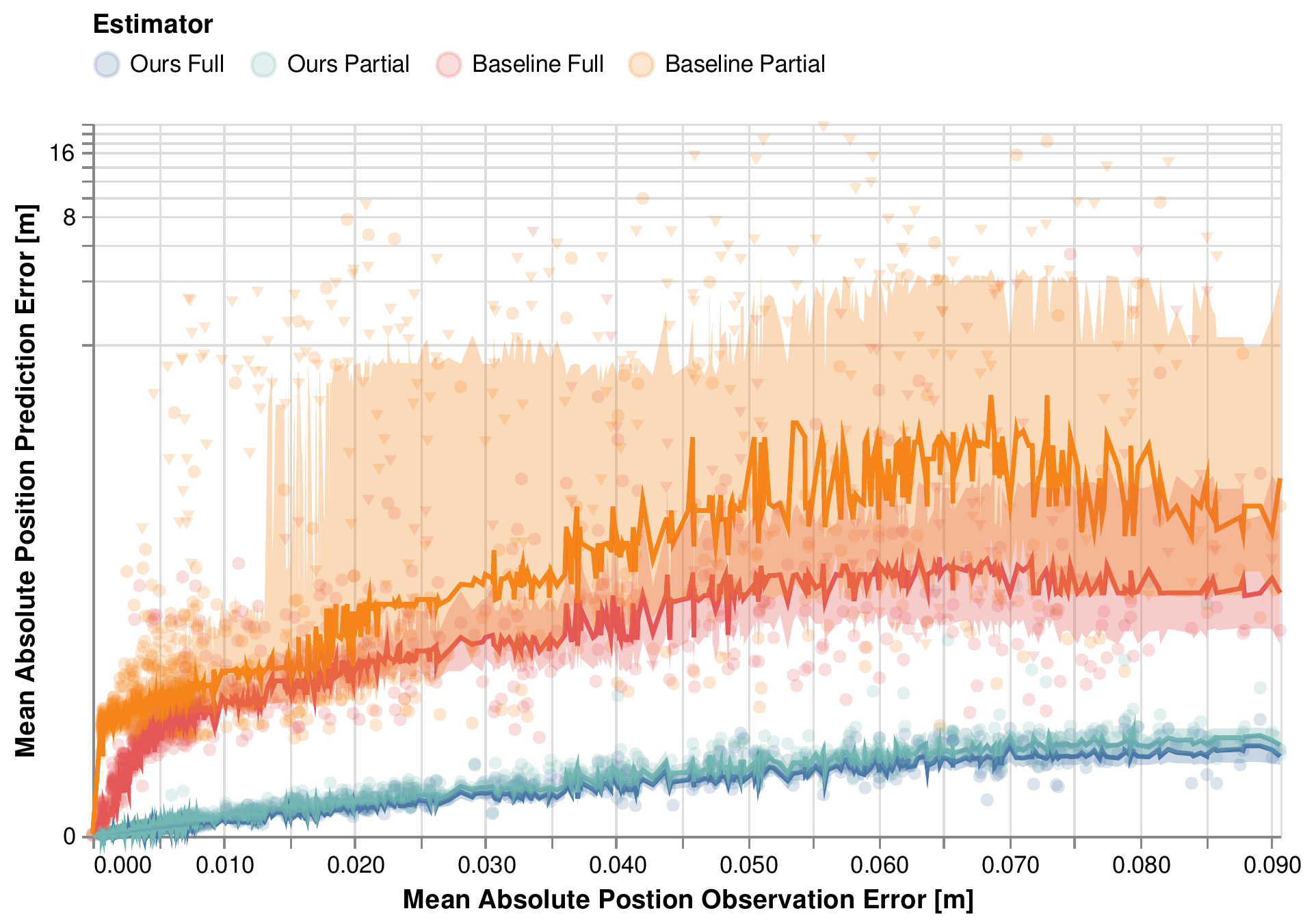}
    }
    \caption{Estimation performance of our method and the baseline for the 2-player running example, with noisy full and partial state observations.
    (a) Error measured directly in parameter space using \cref{eqn:cosine_metric}.
    (b) Error measured in position space.
    Triangular data markers in (b) highlight objective estimates which lead to ill-conditioned games.
    Solid lines and ribbons indicate the median and \ac{iqr} of the error for each case.
    }
    \label{fig:runexmp-estimator_statistics}
\end{figure}

\Cref{fig:runexmp-estimator_statistics} shows the estimator performance for varying levels of observation noise in two different metrics.
\cref{fig:runexmp-estimator_parameter_error} reports the mean cosine error of the objective parameter estimates.
That is, we measure dissimilarity between the unobserved true model parameters $\costparams{}_\textrm{true}$ and the estimate $\costparams{}_\textrm{est}$ by
\begin{align}\label{eqn:cosine_metric}
     D_\textrm{cos}(\costparams{}_\textrm{true}, \costparams{}_\textrm{est}) = 1 -
     \frac{1}{\numplayers} \sum_{i \in [\numplayers]} \frac{\costparams{i\top}_\textrm{true} \costparams{i}_\textrm{est}}{\norm{\costparams{i}_\textrm{true}}_2 \norm{\costparams{i}_\textrm{est}}_2},
\end{align}
where the mean is taken over the $\numplayers$ players.
The normalization of the parameter vectors in \cref{eqn:cosine_metric} reflects the fact that the absolute scaling of the cost weights within each player's objective does not effect their optimal behavior.
Hence, this metric measures the estimator performance in model parameter space.

\begin{figure}
  \centering
  \subfigure[Demonstrations]{
  \includegraphics[scale=\vegascale]{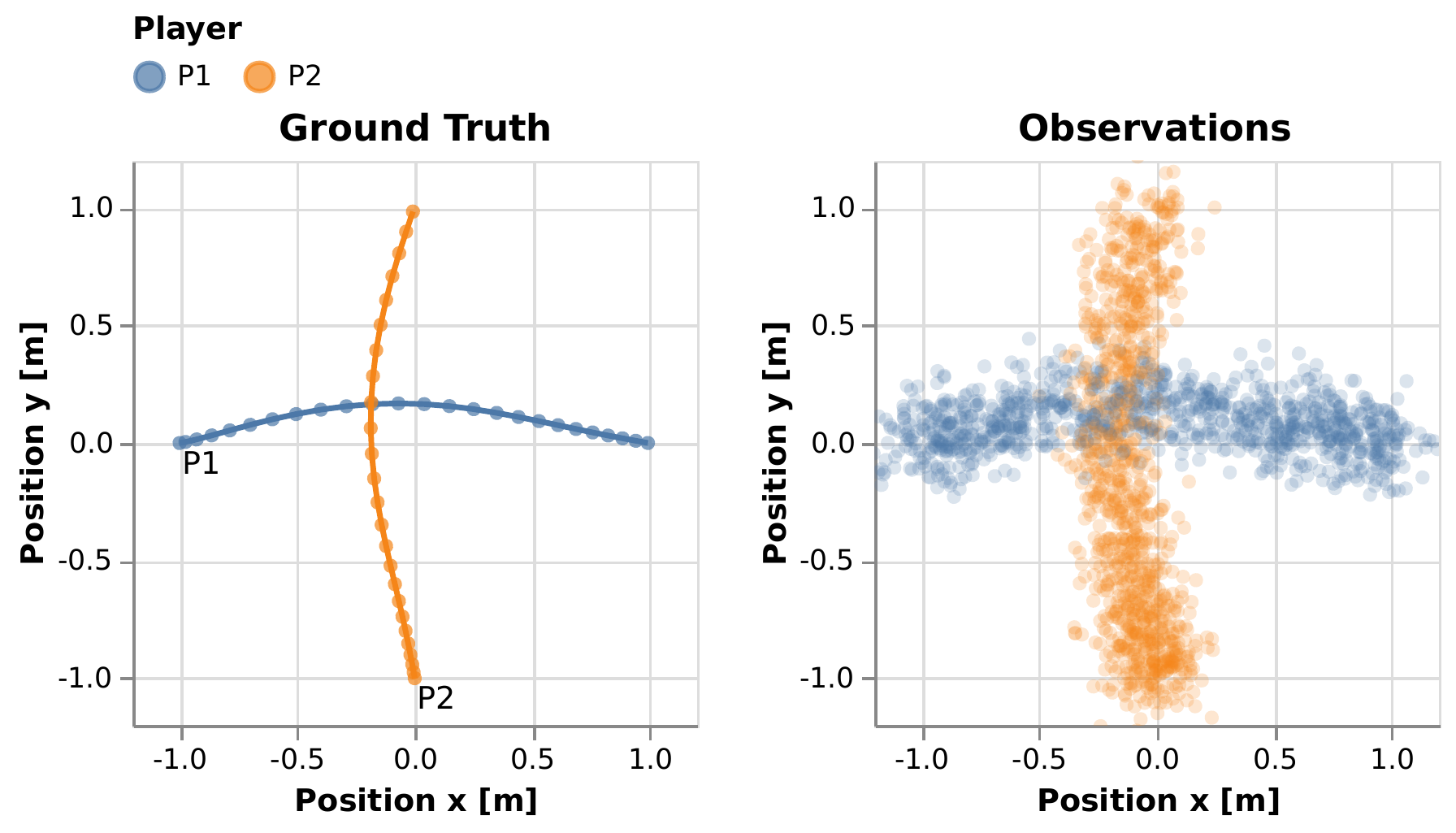}
  \label{fig:runexmp-obs_trajs}
  }
  \hfill
  \subfigure[Ours]{
  \includegraphics[scale=\vegascale]{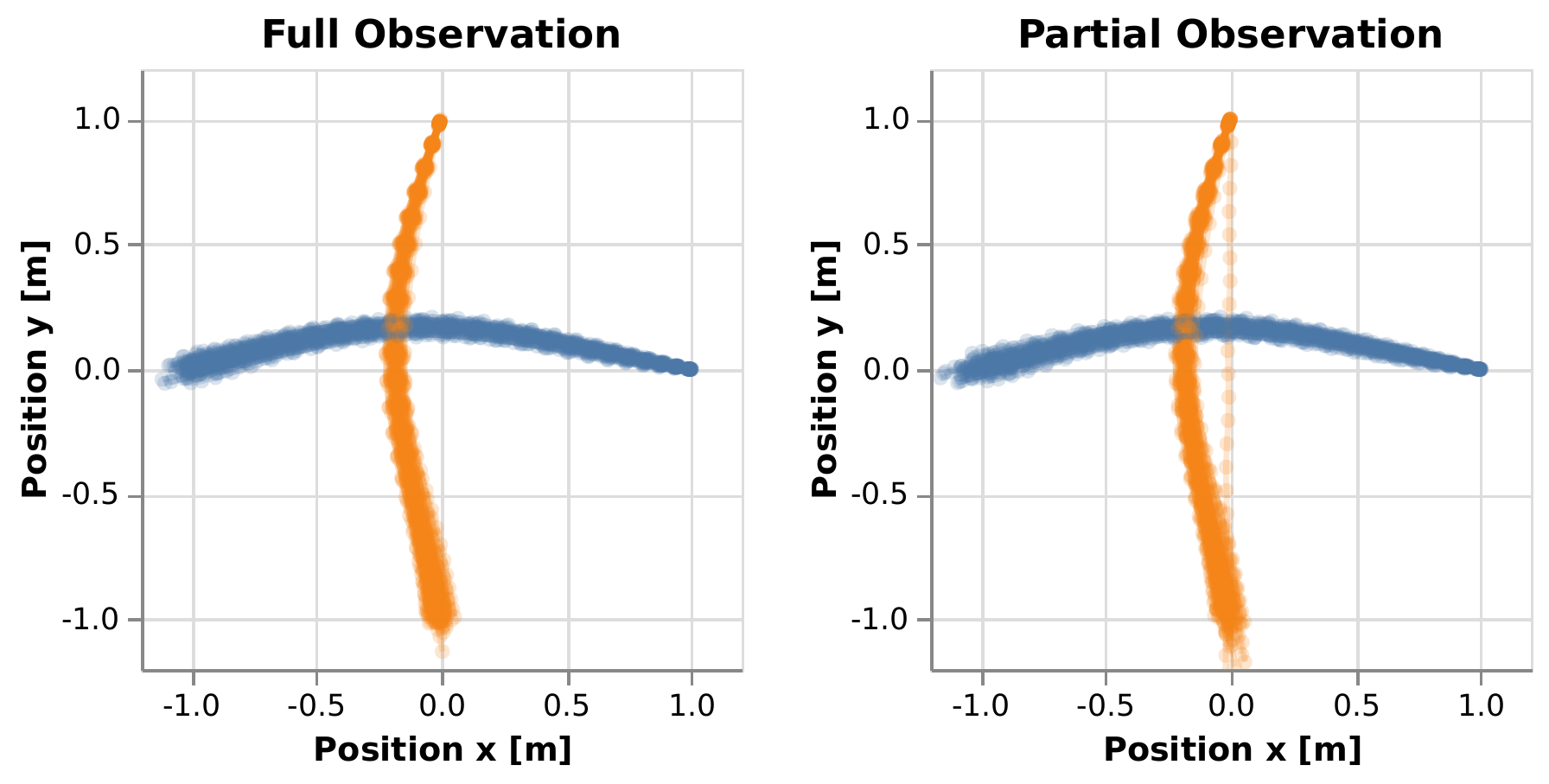}
  \label{fig:runexmp-our_trajs}
  }
  \hfill
  \subfigure[Baseline]{
  \includegraphics[scale=\vegascale]{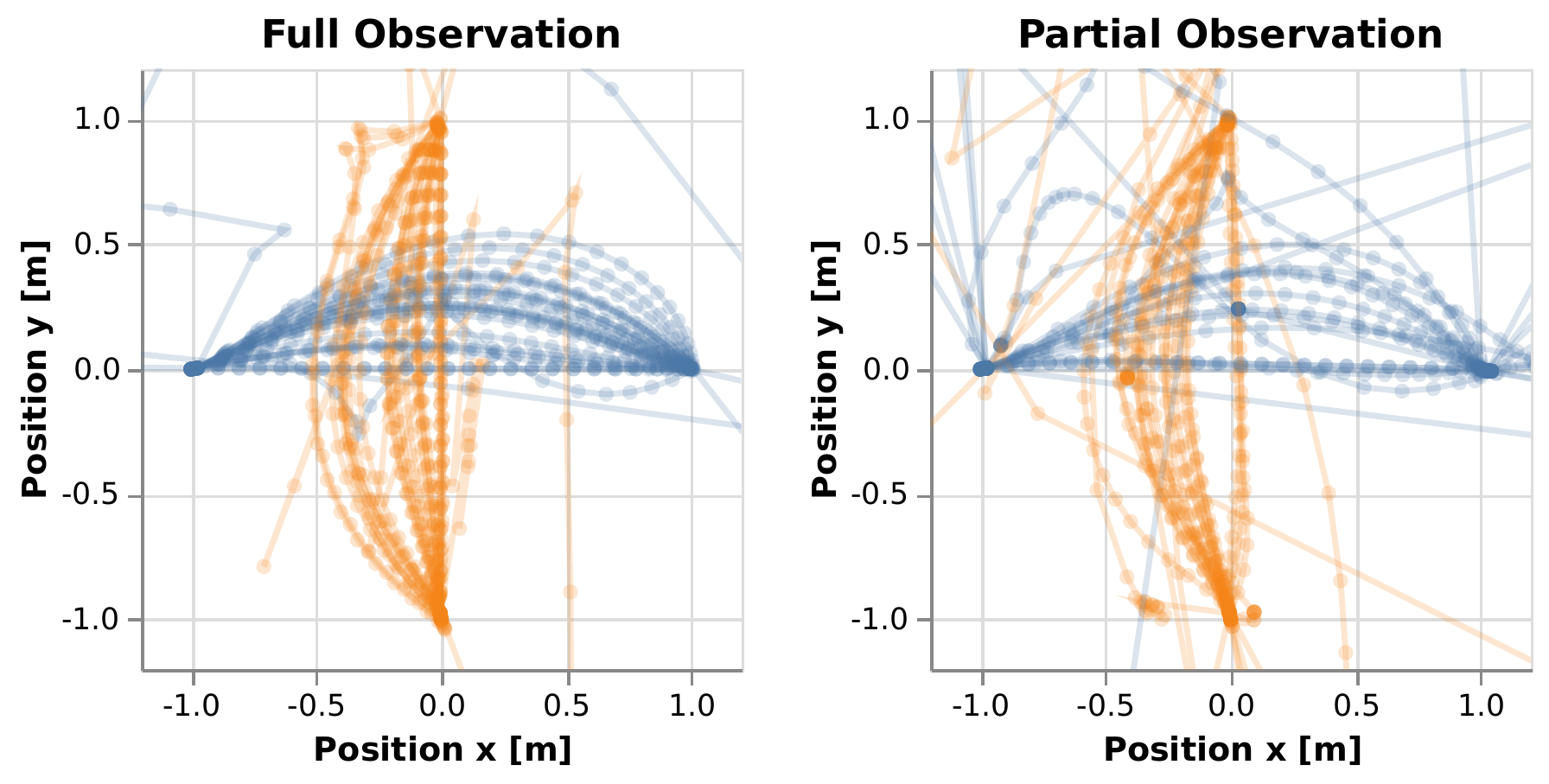}
  \label{fig:runexmp-baseline_trajs}
  }
  \caption{Qualitative prediction performance for the 2-player running example at noise level $\sigma = 0.1$ for 40 different observation sequences. (a) Ground truth trajectory and observations, where each player wishes to reach a goal location opposite their initial position. (b, c) Trajectories recovered by solving the game at the estimated parameters for our method and the baseline using noisy full and partial state observations.
  }
  \label{fig:runexmp-trajs}
\end{figure}

\cref{fig:runexmp-estimator_position_error} shows the mean absolute position error for trajectory predictions computed by finding a root of \cref{eqn:forward_kkt_conditions} using the estimated objective parameters.
We anticipate our method will ultimately be used to forecast the behavior of agents using the estimated objective model, \eg, in a model-predictive control scheme.
For such settings, this metric gives a more tangible sense of algorithmic quality.
In both plots, we evaluate the approaches for two different noisy observation models: in one, estimators observe the \emph{full} state, and in another, estimators observe the position and heading but not the speed of each agent; \ie, they receive a \emph{partial} state observation.
In addition to the raw data, we highlight the median as well as the \ac{iqr} of the estimation error over a rolling window of 60 data points.

\Cref{fig:runexmp-estimator_parameter_error} shows that both methods recover the true cost parameters $\costparams{}$ if observations are not corrupted by noise.
However, the performance of the baseline degrades rapidly with increasing observation noise variance.
This performance degradation is particularly pronounced if the baseline receives only partial state observations.
Our estimator recovers the unknown cost parameters more accurately and with a smaller \ac{iqr}.
In contrast to the baseline, the performance of our method degrades gracefully when observations are corrupted by noise.
This finding also holds if the estimator receives only partial state observations.

\Cref{fig:runexmp-estimator_position_error} displays qualitatively similar patterns, though here the vertical axis measures position prediction error rather than parameter error.
Here, note that some data markers for the baseline estimator are triangles.
These denote instances when the estimated parameters specify ill-conditioned objectives which prevent us from recovering roots of \cref{eqn:forward_kkt_conditions}.
For example, this can happen when proximity costs dominate control input costs.
Thus, these data points correspond to a complete failure of the estimator.
For the baseline, a total of 104 out of 880 estimates result in an ill-conditioned forward game when states are fully observed.
In the case of partial observations, the number of estimator failures increases to 218.
In contrast, our method recovers well-conditioned player objectives for all demonstrations and allows for accurate trajectory prediction.

\begin{figure*}
    \centering
    \subfigure[Parameter estimation\label{fig:highway-estimator_parameter_error}]{
    \includegraphics[scale=\vegascale]{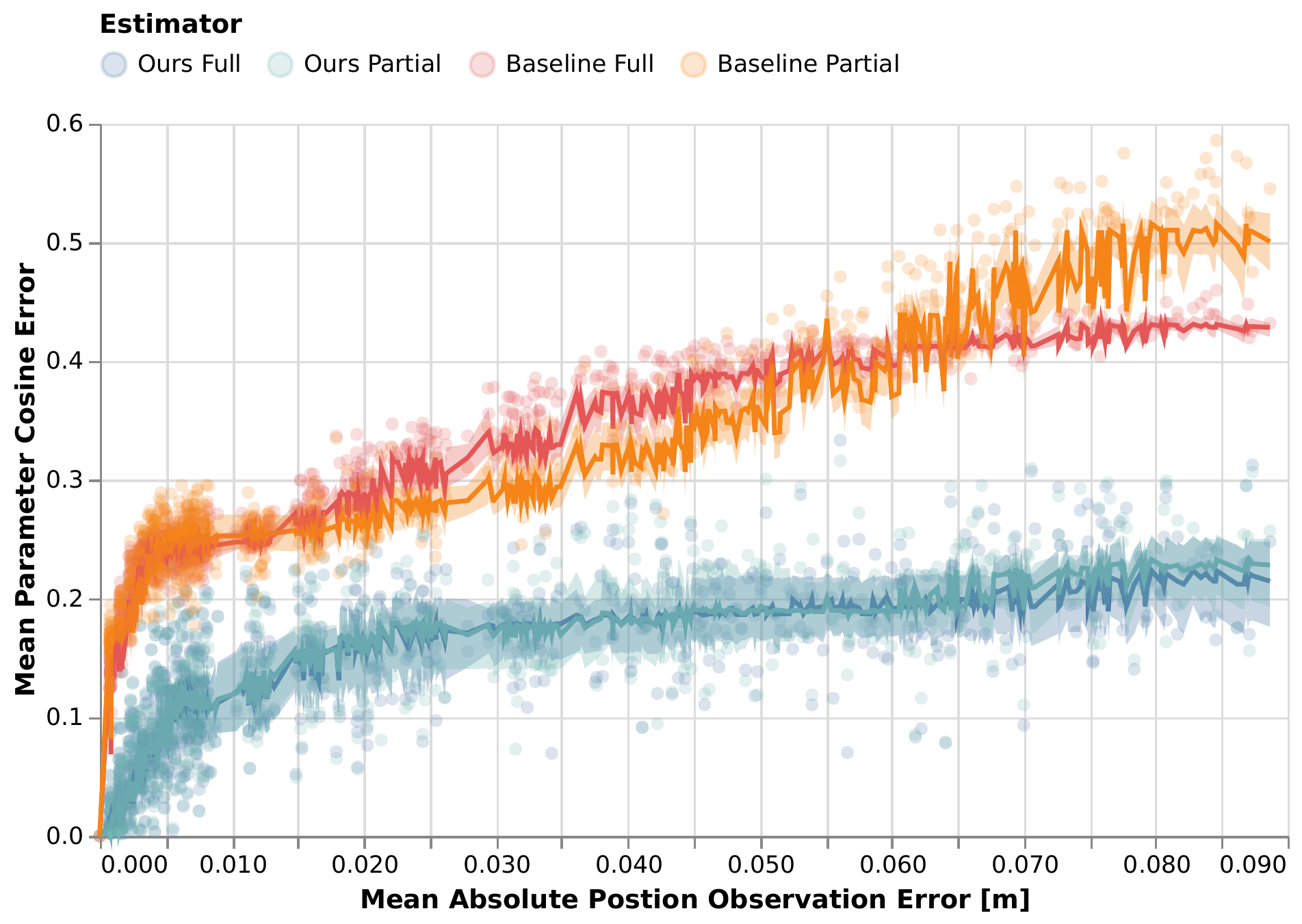}
    }
    \hfill
    \subfigure[Position prediction\label{fig:highway-estimator_position_error}]{
    \includegraphics[scale=\vegascale]{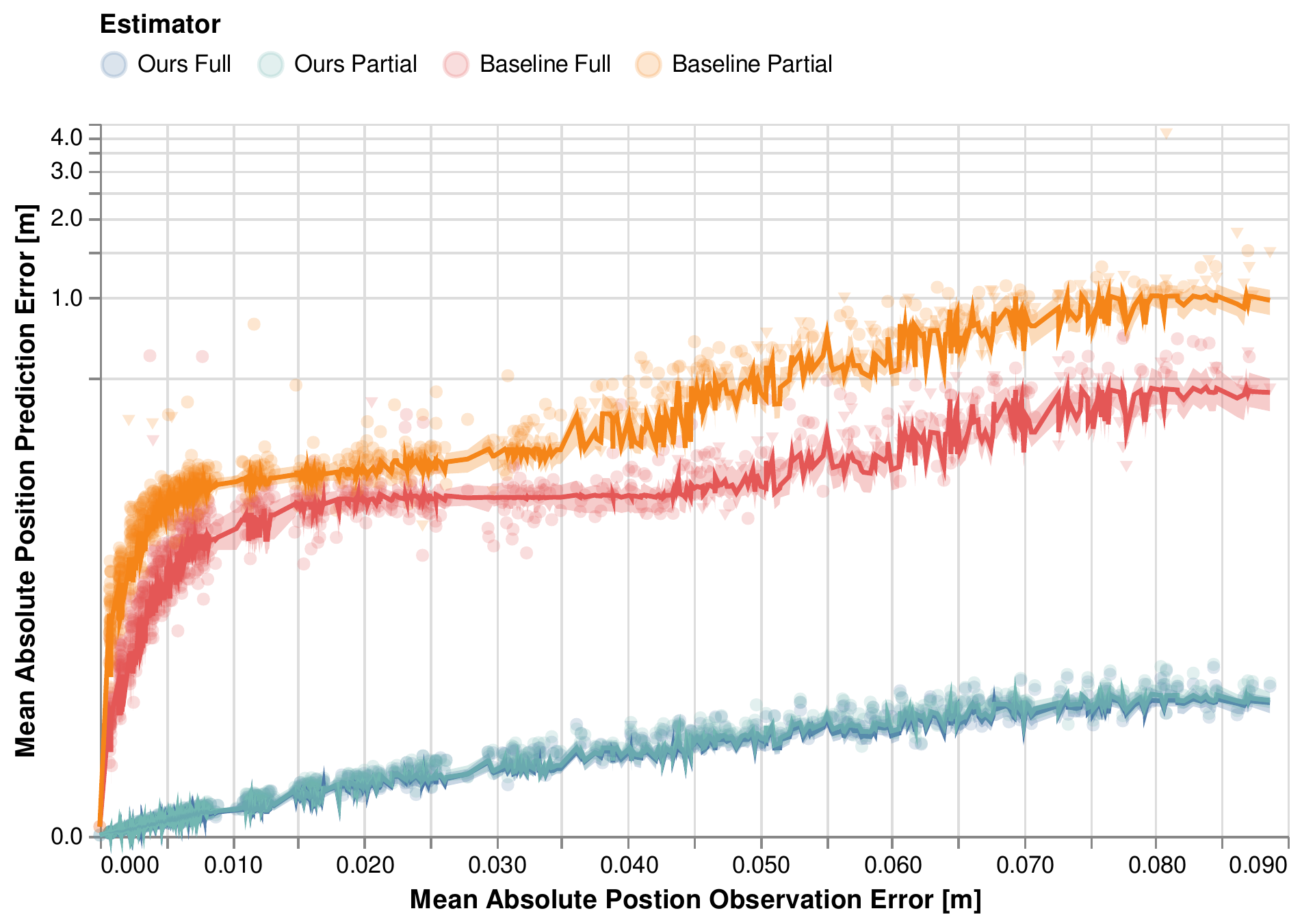}
    }
    \caption{Estimation performance of our method and the baseline for the 5-player highway overtaking example, with noisy full and partial state observations.
    (a) Error measured directly in parameter space using \cref{eqn:cosine_metric}.
    (b) Error measured in position space.
    Triangular data markers in (b) highlight objective estimates which lead to ill-conditioned games.
    Solid lines and ribbons indicate the median and \ac{iqr} of the error for each case.
    \label{fig:highway-estimator_statistics}
    }
\end{figure*}

For additional intuition of the performance gap, \cref{fig:runexmp-trajs} visualizes the prediction results in trajectory space for a fixed initial condition.
\Cref{fig:runexmp-obs_trajs} shows the noise corrupted demonstrations generated for isotropic Gaussian noise with standard deviation $\sigma = 0.1$.
\Cref{fig:runexmp-our_trajs} and \cref{fig:runexmp-baseline_trajs} show the corresponding trajectories predicted by solving the game at the recovered objective estimates of our method and the baseline, respectively.
Note that our method generates a far smaller fraction of outliers than the baseline.
Further, the performance of our method is only marginally effected by partial state observability.

\subsubsection{5-Player Highway Overtaking}
We also replicate the Monte Carlo study in a larger 5-player highway driving scenario in order to demonstrate scalability of the approach.
This scenario is depicted in \cref{fig:frontfig}.
In this highway scenario, each player does not seek to reach a specific goal location at the end of the game horizon.
Instead, a player's objective in this game is more accurately characterized by the desire to make forward progress at an unknown nominal speed.
Therefore, ground-truth objectives use a quadratic penalty on deviation from a desired state that encodes each player's target lane and preferred travel speed.
Note, that this objective can still be modeled by the cost structure in \cref{eqn:runexp-cost}.

\Cref{fig:highway-estimator_statistics} shows the estimator performance of our method and the baseline for this highway driving problem.
Here, we use the same metrics as in the previous experiment to measure estimator performance in parameter space---\cref{fig:highway-estimator_parameter_error}---and position space---\cref{fig:highway-estimator_position_error}.
Again, our method clearly outperforms the baseline for both fully and partially observed demonstrations.
Furthermore, the baseline performance is not consistent across the two metrics.
That is, while the performance of the baseline measured in parameter space is not much effected by partial state observations, the observation model has a decisive impact on the trajectory prediction accuracy.
This performance inconsistency of the baseline can be attributed the fact that certain objective parameters are more critical for accurate prediction of the game trajectory than others.
Since our method's objective is data-fidelity, here measured by observation likelihood \cref{eqn:inverse_objective}, it directly accounts for these effects.
The baseline, however, greedily optimizes the \ac{kkt} residual irrespective of the downstream trajectory prediction task.

\section{Conclusion \& Future Work}
\label{sec:conclusion}

We have proposed a novel method for estimating player objectives from noise-corrupted partial state observations of non-cooperative multi-agent interactions---a task referred to as the \emph{inverse} dynamic game problem.
The proposed solution technique estimates the trajectory to recover unobserved states and inputs, and optimizes this trajectory \emph{simultaneously} with an objective model estimate in order to maximize data-fidelity.
The estimated trajectory is a forward game solution of the observed game including each players' strategy, and may be used for trajectory prediction.

Numerical simulations show that the resulting algorithm is more robust to observation noise and partial state observability than existing methods~\cite{rothfuss2017ifac,awasthi2020acc}, which require estimating states and inputs a priori. Our method recovers model parameters that closely match the unobserved true objectives and accurately predicts the state trajectory; even for high levels of observation noise.

Despite these encouraging results, there is ample room for future improvement.
In the present work, we study the utility of our method for \emph{offline} scenarios in which an external observer recovers the objectives of players \emph{post hoc}.
Our method, however, yields not only the estimated objective model, but also the forward game solution, including each players' strategy.
This property makes our technique particularly suitable for \emph{online} filtering applications in which an autonomous agent must estimate the objectives of other players for safe and efficient closed-loop interaction.
In such a setting, the proposed estimator could be used on a fixed-lag buffer of past observations to simultaneously estimate each opponent's objective while generating the optimal response for the ego-agent over a receding prediction horizon.

Another exciting direction lies in the extension of the proposed method to information structures beyond \ac{olne}. Recent work has put forth efficient solution techniques for the more expressive class of feedback Nash equilibria~\cite{fridovich2020icra,laine2021arxiv}.
While our proposed framework is generally agnostic to the information structure of the observed game, future work should investigate efficient techniques for encoding forward optimality for these equilibrium concepts.


\bibliographystyle{plainnat}
\bibliography{glorified,new}
\end{document}